\documentclass[letterpaper, 10 pt, conference]{ieeeconf}  

\IEEEoverridecommandlockouts                              

\usepackage{graphics} 
\usepackage{epsfig} 
\usepackage{mathptmx} 
\usepackage{times} 
\usepackage{amsmath} 
\usepackage{amssymb}  
\usepackage{cite}
\usepackage{amsfonts}
\usepackage{algorithmic}
\usepackage{graphicx}
\usepackage{textcomp}
\usepackage{xcolor}
\usepackage{makecell}
\usepackage[hidelinks]{hyperref}
\usepackage{bbding}
\usepackage{multirow}
\usepackage{booktabs}
\usepackage{footmisc}
\DeclareMathAlphabet{\mathcal}{OMS}{cmsy}{m}{n}

\title{\LARGE \bf
 Vision-Language Feature Alignment for Road Anomaly Segmentation
}

\author{ $\text{Zhuolin He}$, $\text{Jiacheng Tang}$, $\text{Jian Pu}^{*}$, $\text{Xiangyang Xue}$
\thanks{Z. He, and X. Xue are with the School of Computer Science, Fudan University, Shanghai 200433, China.
J. Tang, and J. Pu are with the Institute of Science and Technology for Brain-Inspired Intelligence, Fudan University, Shanghai, 200433, China}
\thanks{* denotes the corresponding author. (E-mail: jianpu@fudan.edu.cn).}
}

\begin{document}

\maketitle
\thispagestyle{empty}
\pagestyle{empty}

\begin{abstract}

Safe autonomous systems in complex environments require robust road anomaly segmentation to identify unknown obstacles. However, existing approaches often rely on pixel-level statistics to determine whether a region appears anomalous. This reliance leads to high false-positive rates on semantically normal background regions such as sky or vegetation, and poor recall of true Out-of-distribution (OOD) instances, thereby posing safety risks for robotic perception and decision-making. To address these challenges, we propose VL-Anomaly, a vision-language anomaly segmentation framework that incorporates semantic priors from pre-trained Vision-Language Models (VLMs). Specifically, we design a prompt learning-driven alignment module that adapts Mask2Former’s visual features to CLIP text embeddings of known categories, effectively suppressing spurious anomaly responses in background regions. At inference time, we further introduce a multi-source inference strategy that integrates text-guided similarity, CLIP-based image-text similarity and detector confidence, enabling more reliable anomaly prediction by leveraging complementary information sources.
Extensive experiments demonstrate that VL-Anomaly achieves state-of-the-art performance on benchmark datasets including RoadAnomaly, SMIYC and Fishyscapes. Code is released on \url{https://github.com/NickHezhuolin/VL-aligner-Road-anomaly-segment}.

\end{abstract}

\section{INTRODUCTION}

Semantic segmentation plays a pivotal role in enabling autonomous driving systems \cite{ Chen_2023_CVPR,he2025toward, WAM-Flow-2025} and mobile robots \cite{liao2024mobile} to achieve a fine-grained understanding of their surroundings. Typically, segmentation models are trained to recognize a fixed set of pre-defined categories\cite{cheng2022masked,dong2023afformer}. However, in real-world scenarios, these models often encounter out-of-distribution (OOD) objects, such as unexpected obstacles that were not present in the training data. Without proper mechanisms to identify such anomalies, models tend to misclassify these regions into known classes, resulting in inaccurate segmentation masks \cite{blum2021fishyscapes}. Such misclassification undermines model reliability and, more critically, poses substantial safety risks in safety-critical domains such as autonomous driving \cite{bozhinoski2019safety}. Given that fact, accurate segmentation of OOD anomalies is essential for building robust and trustworthy perception systems in open-world environments.

\begin{figure}[t]
\centering
\includegraphics[width=\linewidth]{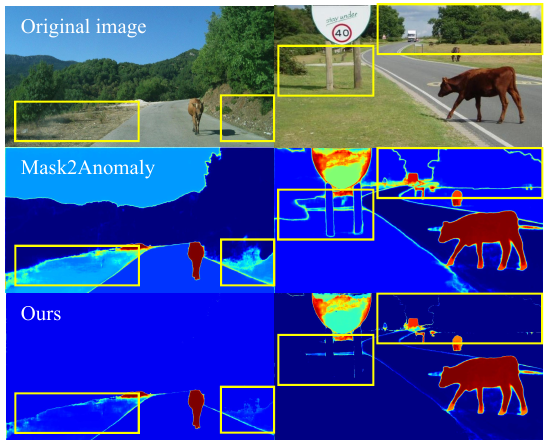}
\caption{Comparison of Anomaly Score Maps. The first row shows the original images, the second row presents anomaly score maps generated by Mask2Anomaly, and the third row illustrates the results of our method. Our approach yields cleaner maps by suppressing false positives on semantically normal background regions such as road surface and vegetation, while more precisely highlighting true anomalies like animals.}
\label{fig:intro}
\end{figure}

Road anomaly segmentation \cite{di2021pixel, grcic2022densehybrid} tackles this challenge by segmenting road objects that fall outside the set of known training classes.
Most existing methods still follow a vision-only paradigm \cite{nayal2023rba,rai2023unmasking}, typically detecting anomalies by thresholding pixel-wise prediction confidence or measuring deviations in low-level visual features. Lacking high-level semantic understanding, these models are constrained to visual similarity, basing their predictions merely on resemblance to feature prototypes or distributions of known classes. Consequently, background regions such as sky and trees belong to normal classes in the dataset, but variations in texture or appearance such as clouds or colors can cause them to be falsely detected as anomalies, as demonstrated in Fig. \ref{fig:intro}. This phenomenon results in a high false-positive rate and undermines the robustness and practicality of existing approaches in real-world autonomous driving scenarios.

Recently, the emergence of vision-language models (VLMs)~\cite{radford2021learning, li2022blip} has introduced a new perspective to road anomaly segmentation. Textual guidance provides explicit semantics, allowing the model to identify pixels corresponding to known categories\cite{jia2021scaling}. At the same time, the semantic dissimilarity derived from vision-language alignment naturally highlights regions that may correspond to unknown objects \cite{ackermann2023maskomaly}. Leveraging semantic similarity not only suppresses false positives but also accentuates rare yet informative OOD cues, thereby enhancing detection accuracy and generalization in open-world scenarios.

Building on vision-language semantic cues, we propose VL-Anomaly, a vision-language anomaly segmentation framework that exploits VLM priors during both training and inference. Its core idea is to integrate category-level knowledge from CLIP~\cite{radford2021learning} into the segmentation process, thereby enhancing the model’s ability to distinguish in-distribution categories from out-of-distribution regions. However, directly applying VLMs to segmentation is non-trivial, as they are not inherently designed for pixel-level multi-class prediction. The core challenge lies in adapting VLMs to the multi-class nature of semantic segmentation. Although recent approaches such as MaskCLIP\cite{dong2023maskclip} demonstrate that multiple categories can be matched within a single sentence, these methods typically rely on handcrafted or concatenated prompts that are not explicitly optimized for segmentation. To overcome this, we design the Prompt Learning-Driven Aligner (PL-Aligner). It employs learnable textual prompts to guide both the backbone features and decoder mask queries into the VLM semantic space. The alignment is carried out at two stages, pixel level and mask level, which together bridge the gap between visual and textual representations. PL-Aligner remains architecture-agnostic and can be seamlessly incorporated into existing segmentation frameworks without requiring structural modifications.

At inference time, we further enhance robustness by adopting a multi-source inference strategy. Specifically, we integrate (i) detector confidence from the segmentation network, (ii) text-guided similarity derived from learned prompts, and (iii) CLIP-based image–text similarity to exploit their complementary strengths. The fusion of these complementary signals provides reliable anomaly predictions and mitigates the weaknesses of relying on a single source. Through this multimodal integration, VL-Anomaly achieves fine-grained anomaly segmentation across diverse road anomaly datasets while maintaining strong generalization. To summarize, we make the following contributions:
\begin{itemize}
\item We propose PL-Aligner, a prompt-driven alignment module that jointly aligns features at both pixel and mask levels, leading to more robust text-guided anomaly segmentation.
\item We introduce a multi-source inference strategy that integrates text-guided similarity, CLIP-based image-text similarity and detector confidence to deliver robust anomaly prediction.
\item Our method delivers consistently state-of-the-art results across RoadAnomaly\cite{lis2019detecting}, Fishyscapes\cite{blum2021fishyscapes} and SMIYC\cite{chan2021segmentmeifyoucan}, showing strong generalization in diverse datasets.
\end{itemize}

\begin{figure*}[t]
\centering
\includegraphics[width=\textwidth]{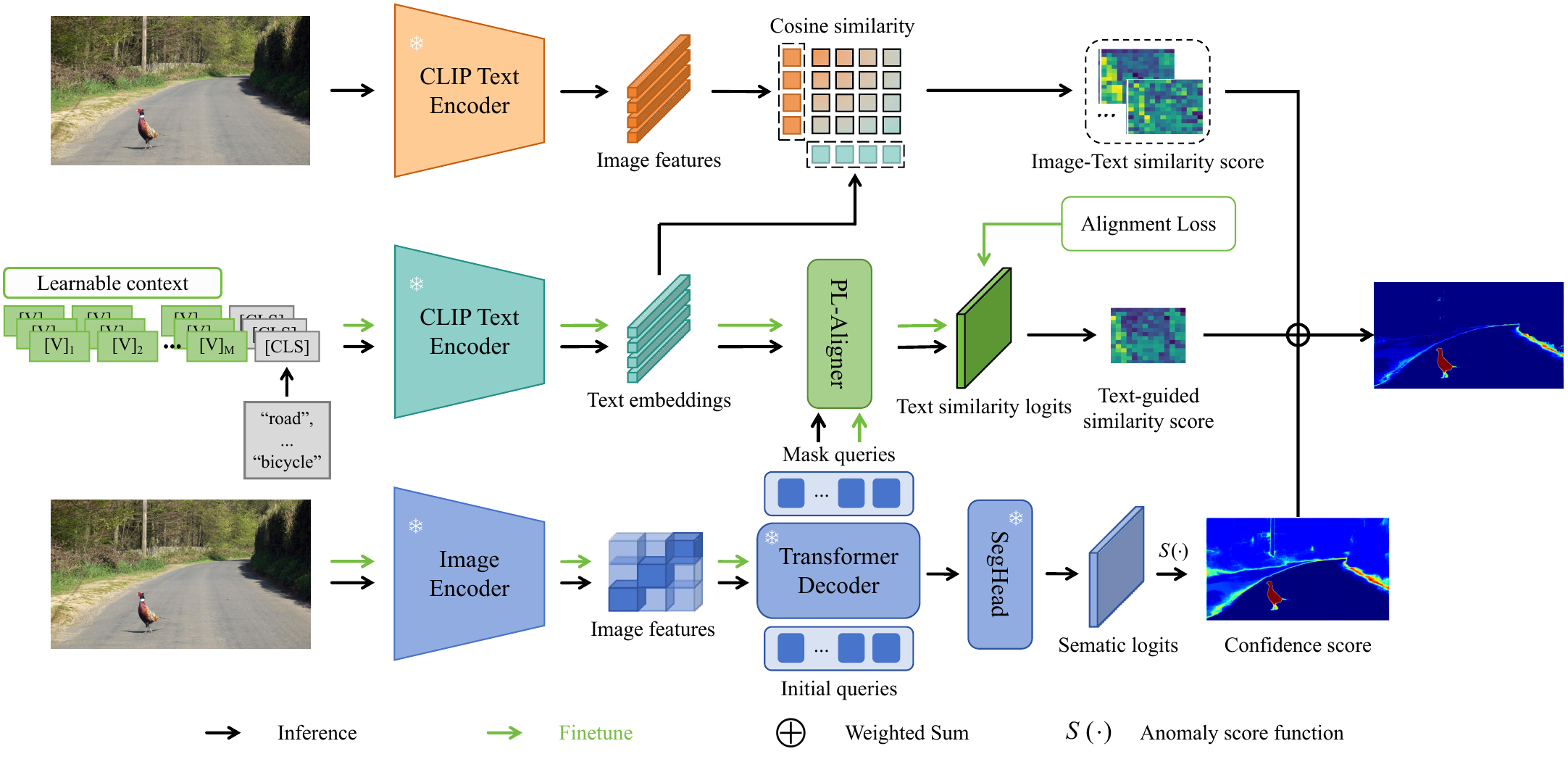}
\caption{Overall architecture of VL-Anomaly. The framework integrates a segmentation backbone with CLIP-based vision–language modules. During training, the Prompt Learning-Driven Aligner (PL-Aligner) first performs pixel-level alignment between the backbone’s visual features and CLIP text embeddings of known categories, and then further establishes mask-level alignment with the decoder’s mask queries. During inference, multi-source scores from the segmentation model outputs, text-guided similarity and CLIP-based image-text similarity are fused to produce robust anomaly segmentation results.}
\label{fig:pipeline}
\end{figure*}

\section{Related Work}
\subsection{Road Anomaly Segmentation}
Road anomaly segmentation aims to localize unexpected obstacles that fall outside the training labels, as evaluated by benchmarks such as Fishyscapes \cite{blum2021fishyscapes} and SMIYC \cite{chan2021segmentmeifyoucan}.

Uncertainty-based methods segment anomalous regions by deriving anomaly scores from softmax/logit statistics or mask-level confidence. MaxLogits \cite{hendrycks2022scaling}, extended from MSP \cite{hendrycks2017a}, has been applied to driving scenarios through the CAOS \cite{hendrycks2019anomalyseg} benchmark. To alleviate miscalibration in confidence scores, SML \cite{jung2021standardized} normalizes the logit distribution and incorporates boundary refinement for better localization of anomalies. Mask2Anomaly \cite{rai2023unmasking} reformulates anomaly segmentation as a mask classification task, achieving improved consistency and recall through mask-level contrastive learning and refinement strategies. Despite these improvements, such methods are prone to false positives, especially in visually distinct but semantically normal regions like trees or the sky.

Outlier exposure-based methods instead leverage auxiliary OOD samples, but their effectiveness depends on the diversity and relevance of the chosen outlier datasets. Maximized Entropy \cite{jung2021standardized} encourages higher-entropy predictions on proxy OOD samples to prevent overconfidence in uncertain regions, while PEBAL \cite{tian2022pixel} employs contrastive learning to differentiate OOD pixels. Similarly, RbA \cite{nayal2023rba} adopts a mask classification paradigm and defines an outlier as any region rejected by all known categories, enabling high-quality OOD segmentation without harming in-distribution performance.

Beyond score-based and outlier-exposure approaches, generative and hybrid methods model inlier feature distributions to obtain likelihood-based OOD scores. DenseHybrid \cite{grcic2022densehybrid} further combines generative modelling of inlier data with discriminative training of negative samples via outlier exposure to enhance OOD segmentation. In addition, image resynthesis-based methods highlight anomalies via reconstruction discrepancies. For example, SynBoost \cite{Di_Biase_2021_CVPR} synthesizes images from semantic maps and combines reconstruction cues with uncertainty-based scores.

In this work, we propose VL-Anomaly, which introduces vision-language priors to suppress false positives. To the best of our knowledge, VL-Anomaly is among the first to incorporate multi-modal semantic priors into \textit{road anomaly segmentation}, offering an alternative solution for robust OOD perception in autonomous driving.

\subsection{Vision-Language Models for Semantic Segmentation}
Vision-language models, such as CLIP \cite{radford2021learning}, have demonstrated impressive generalization in open-world tasks by aligning image and text embeddings in a shared semantic space. Recent efforts have integrated VLMs into semantic segmentation to enhance category extensibility and zero-shot capabilities. Methods like OpenSeg \cite{ghiasi2022scaling} and GroupViT \cite{xu2022groupvit} leverage VLMs to guide pixel or region-level predictions via text supervision, enabling the segmentation of unseen classes. MaskCLIP \cite{dong2023maskclip} further aligns VLM features with segmentation masks to improve visual grounding. However, these works primarily focus on open-vocabulary or zero-shot segmentation within in-distribution domains. More recently, SimCLIP~\cite{deng2024simclip} explored the misalignment between high-level language features and fine-grained visual features in industrial anomaly detection, motivating us to exploit VLM-guided semantic priors for road anomaly segmentation to better capture unseen objects. While most existing approaches \cite{ghiasi2022scaling, xu2022groupvit, dong2023maskclip} perform alignment at a single granularity—either pixel-level or mask-level, CoupAlign \cite{zhang2022coupalign} stands out by coupling the two in a hierarchical manner, which is conceptually closer to our design. 

Unlike open-vocabulary segmentation which uses text prompts as classifiers to expand the label space, our goal is not to recognize novel categories but to leverage VLM priors as semantic regularization for suppressing false positives in anomaly segmentation.

\section{Method}
In this section, we first outline the problem definition, then review a generic mask-transformer architecture in the context of anomaly segmentation, and finally present our proposed framework and its distinctive components, as demonstrated in fig. \ref{fig:pipeline}. 

\subsection{Preliminaries}
Formally, let $\mathcal{X} \subset \mathbb{R}^{3\times H\times W}$ denote the space of RGB images, where $H$ and $W$ are the image height and width, $\mathcal{Y} \subset \mathbb{R}^{C_k\times H\times W}$ denote the space of semantic labels assigning each pixel to one of $C_k$ predefined known categories. Given a training set $\mathcal{D} = \{(x_i, y_i)\}_{i=1}^D$ with $x_i \in \mathcal{X}$ and $y_i \in \mathcal{Y}$, the task of anomaly segmentation is to learn a mapping:  
\begin{equation}
    f: \mathcal{X} \rightarrow \mathbb{R}^{H\times W}
\end{equation}
that produces an anomaly score map for each input image.  

In both per-pixel architectures and mask-transformer architectures, $f(\cdot)$ is ultimately used to assign each pixel or predicted mask, which is then compared against a fixed threshold to distinguish OOD regions from in-distribution (ID) ones. While effective in some cases, such confidence frameworks inherently tie the OOD detection performance to the model’s learned in-distribution representation, making them susceptible to misclassification in unfamiliar scenes. 

To overcome this limitation, we incorporate VLM priors as an external “observer” of semantic knowledge. In contrast to the “insider” perspective of a trained segmentation model, which is confined to task-specific representations, VLMs provide rich open-world semantics that offer an additional viewpoint for distinguishing ID from OOD regions in anomaly segmentation. The following sections detail the technical contributions of our method.

\subsection{Text Prompt Construction}
Unlike typical VLM applications \cite{radford2021learning} that process a single category description at a time, semantic segmentation requires handling multiple categories simultaneously, as a single image often contains several semantic classes. Compared with manually crafted natural language sentences, adopting the learnable prompt paradigm not only avoids ambiguity and redundancy but also enables automatic adaptation to the segmentation task through joint training, thereby achieving more robust cross-modal alignment~\cite{zhou2022coop}.

To enable parallel alignment with all known categories, we construct a dedicated prompt for each class $c_i \in \mathcal{C}$, where $\mathcal{C} = \{c_1, c_2, \dots, c_{C_k}\}$ denotes the set of $C_k$ known semantic classes (e.g., $C_k=19$ for Cityscapes\cite{cordts2016cityscapes}). 
This design allows the model to compute class-wise vision-language similarities in a single forward pass, enabling efficient dense prediction.
Following the learnable prompt paradigm~\cite{zhou2022coop}, each prompt adopts a unified context form:
\begin{equation}
    \mathbf{p}_i = [\mathbf{V}]_1\,[\mathbf{V}]_2\,\dots\,[\mathbf{V}]_{M}\,[\text{CLS}],
\end{equation}
where $[\mathbf{V}]_i$ ($i \in \{1, \dots, M\}$) denotes a learnable context token with the same dimension $d$ as the VLM word embeddings, $M$ is the number of context tokens, and $[\text{CLS}]$ is the textual name of class $c_i$. 
The context tokens share the same form across categories and are optimized jointly with the segmentation model to capture task-specific semantics while preserving the VLM’s open-world prior.

Given the constructed prompt $\mathbf{p}_i$, the category-specific text embedding is obtained via the VLM’s text encoder:
\begin{equation}
    \mathbf{t}_i = \mathrm{TextEncoder}(\mathbf{p}_i),
\end{equation}
where $\mathbf{t}_i \in \mathbb{R}^{d}$ denotes the text embedding for class $c_i$.

\subsection{Prompt Learning-Driven Aligner}
\begin{figure}[t]
\centering
\includegraphics[width=\linewidth]{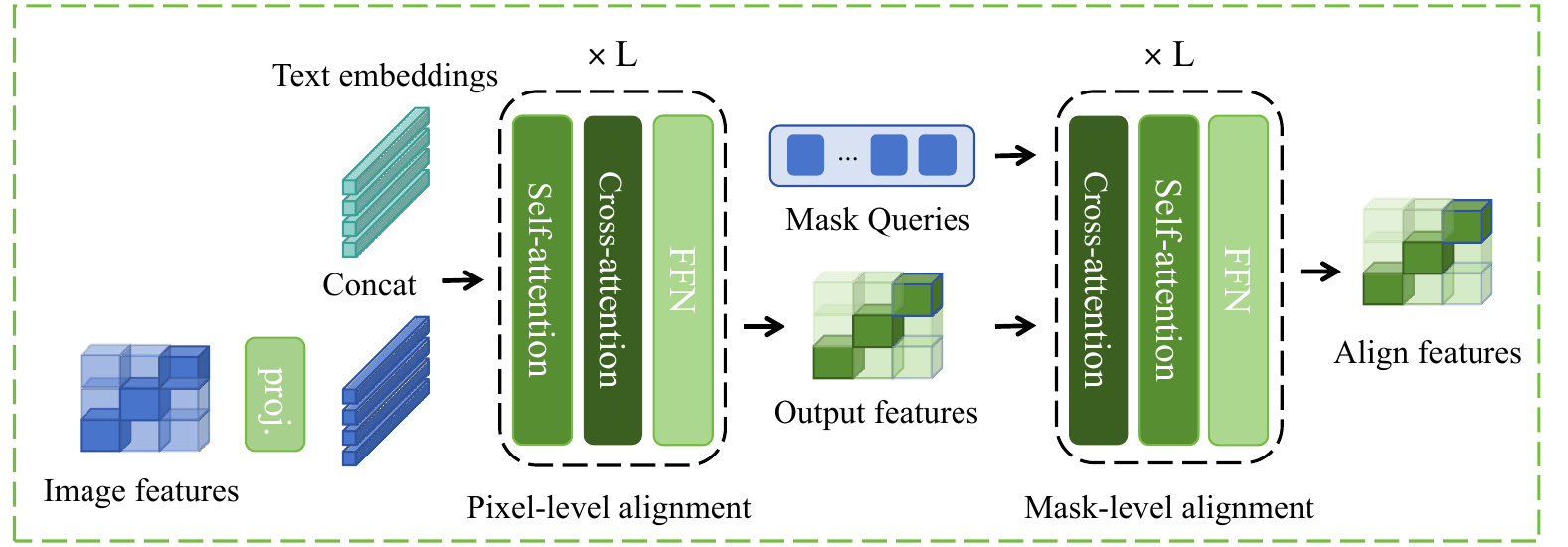}
\caption{Architecture of PL-Aligner. The first layer aligns pixel-level visual features from the backbone with text embeddings, while the second layer aligns mask queries from the decoder with the pixel-level features from the first layer to achieve mask-level alignment. Standard operations such as normalization and activation functions are omitted for clarity.}
\label{fig:PL-Aligner}
\end{figure}

The feature spaces of the segmentation model and the VLM are inherently misaligned, motivating our learnable prompt-based alignment mechanism. Existing VLM-based segmentation approaches typically adopt a single-granularity strategy, such as pixel-level feature alignment \cite{jia2021scaling} or mask-level alignment based on region features \cite{dong2023maskclip}. Inspired by CoupAlign~\cite{zhang2022coupalign}, which demonstrates the benefit of combining pixel- and mask-level alignments, we propose a PL-Aligner that jointly enforces fine-grained pixel consistency and structured mask-level semantic alignment, see fig.\ref{fig:PL-Aligner}. 

\paragraph{Pixel-level alignment}
For an input image, the visual encoder first generates a dense feature map, where each spatial location $i$ corresponds to a feature vector $\mathbf{v}_i$. To align these features with the text prompts, the image feature map is re-projected into the same embedding dimension as the text embeddings, concatenated with the prompt features, and processed jointly through attention mechanisms and feed-forward networks. The resulting pixel-aligned features are then aligned with the corresponding text embeddings $\mathbf{t}_{y_i}$ using a pixel-level contrastive loss:
\begin{equation}
\mathcal{L}_{\mathrm{pixel}} = - \frac{1}{N} \sum_{i=1}^{N} 
\log \frac{\exp\left(\mathrm{sim}(\mathbf{v}_i, \mathbf{t}_{y_i}) / \tau \right)}
{\sum_{k=1}^{C_k} \exp\left(\mathrm{sim}(\mathbf{v}_i, \mathbf{t}_k) / \tau \right)} ,
\end{equation}
where $\mathrm{sim}(\cdot,\cdot)$ denotes cosine similarity, $\tau$ is a learnable temperature parameter, and $y_i$ is the ground-truth label. This stage enforces direct alignment between pixel-level visual features and textual semantics.  

\paragraph{Mask-level alignment}
Building upon the pixel-aligned features, we further introduce mask-level alignment after the mask-transformer decoder. Specifically, the pixel-aligned features are used as keys and values, while the decoder mask queries serve as queries. Through attention mechanisms and feed-forward networks, the decoder queries are refined into semantically consistent representations, and then aligned with the text embeddings in the same contrastive manner:  
\begin{equation}
\mathcal{L}_{\mathrm{mask}} = - \frac{1}{N_q} \sum_{m=1}^{N_{q}} 
\log \frac{\exp\left(\mathrm{sim}(\mathbf{q}_m, \mathbf{t}_{y_m}) / \tau \right)}
{\sum_{k=1}^{C_k} \exp\left(\mathrm{sim}(\mathbf{q}_m, \mathbf{t}_k) / \tau \right)} ,
\end{equation}
where $\mathbf{q}_m$ denotes the projected mask query and $y_m$ is the ground-truth class label for mask $m$. This stage ensures that the mask-level representations are not only consistent with the pixel-level aligned features but also aligned with the semantic space of the VLM. 

\paragraph{Alignment loss}
The final alignment loss combines the segmentation supervision with the two alignment objectives:  
\begin{equation}
\mathcal{L}_{\mathrm{align}} = \mathcal{L}_{\mathrm{seg}} 
+ \lambda_{\mathrm{pixel}} \mathcal{L}_{\mathrm{pixel}} 
+ \lambda_{\mathrm{mask}} \mathcal{L}_{\mathrm{mask}} ,
\end{equation}
where $\mathcal{L}_{\mathrm{seg}}$ represents the standard segmentation loss following Mask2Former~\cite{rai2023unmasking}, with $\lambda_{\mathrm{pixel}}$ and $\lambda_{\mathrm{mask}}$ fixed to 0.5 in our finetuning stage.

\begin{table*}[t]
\begin{center}
    \caption{Evaluation on RoadAnomaly \cite{lis2019detecting}, SMIYC-RA21 \cite{chan2021segmentmeifyoucan} and SMIYC-RO21 \cite{chan2021segmentmeifyoucan}.}
    \label{tab-ra-smiyc}
    \scriptsize
    \setlength{\tabcolsep}{5.6pt} 
    \begin{tabular}{l|ccc|ccc|ccc|ccc}
    \toprule[1.5pt]
        \multirow{2}{*}{\textbf{Methods}} 
        & \multicolumn{3}{c|}{\textbf{RoadAnomaly}}  
        & \multicolumn{3}{c|}{\textbf{SMIYC-RA21}} 
        & \multicolumn{3}{c|}{\textbf{SMIYC-RO21}}
        & \multicolumn{3}{c}{\textbf{Average}} \\
        ~ & FPR$_{95}$$\downarrow$ & AuPRC$\uparrow$ & AuROC$\uparrow$ 
           & FPR$_{95}$$\downarrow$ & AuPRC$\uparrow$ & AuROC$\uparrow$ 
           & FPR$_{95}$$\downarrow$ & AuPRC$\uparrow$ & AuROC$\uparrow$
           & FPR$_{95}$$\downarrow$ & AuPRC$\uparrow$ & AuROC$\uparrow$ \\
        \midrule[1.pt]
        MSP \cite{hendrycks2017a} & 71.4 & 15.7 & - & 72.0 & 28.0 & - & 16.6 & 15.7 & - & 53.3 & 19.8 & - \\
        Entropy \cite{hendrycks2017a} & 68.2 & 15.7 & - & - & - & - & - & - & - & 68.2 & 15.7 & - \\
        Mahalanobis \cite{lee2018simple} & 81.1 & 14.4 & - & 87.0 & 20.0 & - & 13.1 & 20.1 & - & 60.4 & 18.2 & - \\
        SML \cite{jung2021standardized} & 70.7 & 17.5 & - & 39.5 & 46.8 & - & 36.8 & 3.4 & - & 49.0 & 22.6 & - \\
        JSRNet \cite{vojir2021road} & \textbf{9.2} & \textbf{94.4} & - & 43.9 & 33.6 & - & 28.9 & 28.1 & - & 27.3 & 52.0 & - \\
        SynBoost \cite{di2021pixel} & 64.8 & 38.2 & - & 61.9 & 56.4 & - & 3.2 & 71.3 & - & 43.3 & 55.3 & - \\
        Max Entropy \cite{chan2021entropy} & 31.8 & 48.9 & - & 15.0 & 85.5 & - & 0.8 & 85.1 & - & 15.9 & 73.2 & - \\
        Dense Hybrid \cite{grcic2022densehybrid} & 64.0 & 31.4 & - & 9.8 & 78.0 & - & \textbf{0.2} & 87.1 & - & 24.7 & 65.5 & - \\
        PEBEL \cite{tian2022pixel} & 44.6 & 45.1 & - & 40.8 & 49.1 & - & 12.7 & 5.0 & - & 32.7 & 33.1 & - \\
        \midrule[1.pt]
        ODIN$^\dagger$ \cite{liang2018enhancing} & 32.9 & 55.0 & 91.4 & \underline{3.6} & 91.5 & 98.5 & 1.1 & 47.8 & \underline{99.2} & 12.5 & 64.8 & 96.4 \\
        Mask2Anomaly$^\dagger$ \cite{rai2023unmasking} & 13.2 & \underline{79.7} & \underline{96.2} & 14.6 & \underline{88.7} & \textbf{99.7} & \textbf{0.2} & \textbf{93.3} & 99.1 & \underline{9.3} & \underline{87.2} & \underline{98.3} \\
        VL-Anomaly (Ours) & \underline{12.9} & 79.2 & \textbf{96.8} & \textbf{3.5} & \textbf{95.1} & \textbf{99.7} & 0.6 & \underline{91.0} & \textbf{99.7} & \textbf{5.7} & \textbf{88.4} & \textbf{98.7} \\
        \bottomrule[1.5pt]
        \multicolumn{13}{p{.96\textwidth}}{
            \textit{Note:} $\uparrow$ indicates higher is better, $\downarrow$ indicates lower is better. The best and second best results are \textbf{bold} and \underline{underlined}, respectively.
        }
    \end{tabular}
\end{center}
\end{table*}

\subsection{Multi-source Inference Strategy}
Building on the previous section, where the PL-Aligner introduces text awareness during training, we further incorporate external semantic priors from CLIP at inference to better separate ID from OOD regions.
To this end, we propose a multi-source inference strategy that integrates three complementary scores: (i) detector confidence, (ii) text-guided similarity, and (iii) CLIP-based image–text similarity.
This strategy mitigates the weaknesses of relying on a single source.

\paragraph{Detector confidence}
Inspired by \cite{rai2023unmasking}, for a mask-transformer architecture, the decoder outputs class scores $\mathbf{c}_m \in \mathbb{R}^{C_k}$ and mask logits $\mathbf{m}_m \in \mathbb{R}^{H\times W}$ for each mask index $m$. The detector confidence map for class $k$ is computed as:
\begin{equation}
    S_{\mathrm{conf}} = \max_{k \in \{1,\dots,C_k\}} \mathrm{softmax}(\mathbf{c}_m)_k \cdot \sigma(\mathbf{m}_m),
\end{equation}
where $\sigma(\cdot)$ denotes the element-wise sigmoid function.

\paragraph{Text-guided similarity}
We reuse the learned prompt embeddings $\mathbf{t}_k$ obtained during training. The final align features $\mathbf{v}_{align}$ from the PL-Aligner are compared with $\mathbf{t}_k$ to compute the text-guided similarity score:
\begin{equation}
    S_{\mathrm{text}}^{(k)} = \mathrm{sim}(\mathbf{v}_{align}, \mathbf{t}_k).
\end{equation}
This score quantifies the semantic consistency between the aligned visual representation and each learned class-specific prompt embedding.

\paragraph{CLIP-based image-text similarity}
We also compute an image-level similarity score between the input image $x$ and each class prompt $\mathbf{t}_k$ using the frozen CLIP image encoder:
\begin{equation}
    S_{\mathrm{img}}^{(k)} = \mathrm{sim}(\mathrm{ImageEncoder}(x), \mathbf{t}_k),
\end{equation}
which provides a global semantic prior that is independent of the segmentation model’s predictions.

\paragraph{Score Fusion}
The three scores are combined into a unified anomaly score:
\begin{equation}
    S_{\mathrm{final}} = 1 - \max_{k \in \{1,\dots,C_k\}} \Big( 
    \alpha \cdot S_{\mathrm{conf}}^{(k)} +
    \beta \cdot S_{\mathrm{text}}^{(k)} +
    \gamma \cdot S_{\mathrm{img}}^{(k)} 
    \Big),
\end{equation}
where $\alpha$ ,$\beta$, $\gamma$ are set to 0.7, 0.2 and 0.1, respectively. A higher $S_{\mathrm{final}}$ indicates a greater likelihood of an OOD region.

\section{Experiments}
In this section, we first introduce the experiment setup, including the introduction of the anomaly inference datasets, the evaluation metrics and the implementation details. Then, we compare our method with other outstanding baselines. Besides, we provide extensive ablations on both VL-Anomaly modules and the layer design of PL-Aligner.

\subsection{Setup}
\noindent\textbf{Datasets.}\quad
The anomaly inference model is first finetuned on Cityscapes \cite{cordts2016cityscapes} (2,975 training, 500 validation and 1,525 test images), then fine-tuned following \cite{rai2023unmasking} using outlier auxiliary datasets generated from MS-COCO \cite{lin2014microsoft}.
We evaluate on three benchmarks: RoadAnomaly \cite{lis2019detecting}, Segment Me If You Can (SMIYC) \cite{chan2021segmentmeifyoucan} and Fishyscapes \cite{blum2021fishyscapes}. RoadAnomaly contains 60 Internet-sourced validation images with anomalies of varying scales. SMIYC includes two subsets: RoadAnomaly21 (RA21, 10 validation and 100 test images) and RoadObstacle21 (RO21, 30 validation and 327 test images). Fishyscapes has two subsets: Static (30 validation and 1,000 test images) and Lost \& Found (L\&F, 100 validation and 275 test images).
During inference, only the validation split ground truth is accessible for RoadAnomaly, RA21, RO21 and Fishyscapes; the test ground truth remains unavailable.

\noindent\textbf{Evaluation metrics.}\quad 
Following \cite{tian2022pixel, liu2023residual, rai2023unmasking}, we report the Area Under the Receiver Operating Characteristic curve (AuROC), the False Positive Rate at 95\% true positive rate ($\text{FPR}_{95}$) and the Area Under the Precision-Recall Curve (AuPRC) for pixel-level evaluation. 
However, pixel-level metrics may overlook small anomalies and be biased toward large anomalous regions. 
Therefore, we additionally adopt component-level metrics, including the averaged component-wise F1 score ($\text{F1}^*$), the Positive Predictive Value (PPV) and the component-wise Intersection over Union (sIoU).

\noindent\textbf{Baselines.}\quad In addition to comparing with the results reported in prior works \cite{hendrycks2017a, chan2021entropy,lee2018simple,vojir2021road,jung2021standardized,di2021pixel,grcic2022densehybrid,tian2022pixel}, we further reproduced ODIN$^\dagger$ \cite{liang2018enhancing} and Mask2Anomaly$^\dagger$ \cite{rai2023unmasking} for a fairer and more comprehensive comparison. For ODIN \cite{liang2018enhancing}, the temperature parameter was set to $T=3.0$, following the implementation provided in the SMIYC benchmark \cite{chan2021segmentmeifyoucan} repository. For Mask2Anomaly \cite{rai2023unmasking}, the refine mask stage was excluded during training. Since the former is a classical work on confidence-based OOD detection and the latter serves as our main baseline for mask-level anomaly segmentation, their inclusion highlights the rationality of our comparison. It is worth noting that all reproduced baselines were applied without any modification to the Mask2Former model architecture, ensuring the fairness of the comparison.

\noindent\textbf{Implementation details.}
Our mask segmentation network is based on Mask2Anomaly~\cite{rai2023unmasking}, which builds upon Mask2Former~\cite{cheng2022masked} with a ResNet-50~\cite{he2016deep} backbone. During finetuning, the segmentation network and CLIP ViT-L-16 weights are frozen. By default, we employ L = 1 layers of pixel-level and mask-level attention. For prompt learning, we adopt $M=16$ learnable context tokens per class, each of dimension $d=512$, initialized from predefined templates via the CLIP’s word embeddings, similar to MaskCLIP\cite{dong2023maskclip}.
The segmentation loss $\mathcal{L}_{\mathrm{seg}}$ follows the standard configuration in Mask2Former, the loss is a weighted sum of classification, Dice\cite{milletari2016v} and mask losses with coefficients ${1.0, 1.0, 20.0}$, respectively. The temperature parameter $\tau$ in the contrastive loss is treated as a learnable scalar, initialized to 0.07 following common practice in contrastive learning~\cite{radford2021learning}.
For optimization, the model is finetuned using AdamW~\cite{loshchilov2017decoupled} with an initial learning rate of $1\text{e}{-4}$, weight decay 0.05 and batch size 8 for 5,000 iterations. 
The training images are randomly cropped to $380\times760$, and a random scale sampled from from 0.1 to 2.0.
During validation, we compute anomaly scores for ID and OOD pixels separately using ground-truth OOD masks, consistent with prior works~\cite{rai2023unmasking}. All experiments are conducted on a single NVIDIA 3090 GPU.

\begin{table}[ht]
\begin{center}
    \caption{Evaluation on Fishyscapes Static and Fishyscapes Lost \& Found validation datasets.}
    \label{tab-fs}
    \scriptsize
    \setlength{\tabcolsep}{2.pt} 
    \begin{tabular}{l|ccc|ccc}
    \toprule[1.5pt]
        \multirow{2}{*}{\textbf{Methods}} 
        & \multicolumn{3}{c|}{\textbf{FS Static}} 
        & \multicolumn{3}{c}{\textbf{FS L\&F}} \\
        ~ & FPR$_{95}$$\downarrow$ & AuPRC$\uparrow$ & AuROC$\uparrow$ 
           & FPR$_{95}$$\downarrow$ & AuPRC$\uparrow$ & AuROC$\uparrow$ \\
        \midrule[1.pt]
        MSP \cite{hendrycks2017a} & 39.8 & 12.9 & - & 44.8 & 1.3 & - \\
        Entropy \cite{hendrycks2017a} & 39.7 & 15.4 & - & 44.8 & 2.9 & - \\
        SML \cite{jung2021standardized} & 20.5 & 52.1 & - & 21.9 & 31.7 & - \\
        SynBoost \cite{di2021pixel} & 15.8 & 43.2 & - & 18.8 & \textbf{72.6} & - \\
        Max Entropy \cite{chan2021entropy} & 8.6 & 86.5 & - & 35.1 & 29.9 & - \\
        Dense Hybrid \cite{grcic2022densehybrid} & 5.9 & 80.2 & - & 3.9 & 47.1 & - \\
        PEBEL \cite{tian2022pixel} & \textbf{1.7} & \textbf{92.4} & - & 7.6 & 44.2 & - \\
        \midrule[1.pt]
        ODIN$^\dagger$ \cite{liang2018enhancing} & 62.4 & 82.3 & 89.4 & 81.1 & 1.4 & 86.3 \\
        Mask2Anomaly$^\dagger$ \cite{rai2023unmasking} & \underline{1.9} & 90.4 & \underline{98.3} & \textbf{4.4} & 46.0 & \underline{93.6} \\
        VL-Anomaly (Ours) & 2.3 & \underline{90.7} & \textbf{98.7} & \underline{8.4} & \underline{69.5} & \textbf{96.0} \\
        \bottomrule[1.5pt]
        \multicolumn{7}{p{.96\linewidth}}{
            \textit{Note:} $\uparrow$ indicates higher is better, $\downarrow$ indicates lower is better. The best and second best results are \textbf{bold} and \underline{underlined}, respectively.
        }
    \end{tabular}
\end{center}
\end{table}

\begin{table}[ht]
\centering
\caption{Results on the SMIYC \cite{chan2021segmentmeifyoucan} benchmark.}
\label{tab-smiyc}
\scriptsize
\setlength{\tabcolsep}{5.5pt} 
\begin{tabular}{l|ccc|ccc}
\toprule[1.5pt]
\multirow{2}{*}{\textbf{Methods}} 
& \multicolumn{3}{c|}{\textbf{SMIYC-RA21}} 
& \multicolumn{3}{c}{\textbf{SMIYC-RO21}} \\
~ & \textbf{sIoU$\uparrow$} & \textbf{PPV$\uparrow$} & \textbf{F1$^*\uparrow$} 
   & \textbf{sIoU$\uparrow$} & \textbf{PPV$\uparrow$} & \textbf{F1$^*\uparrow$} \\
\midrule[1.pt]
MSP \cite{hendrycks2017a}         & 15.5 & 15.3 &  5.4 & 19.7 & 15.9 &  6.3 \\
Mahalanobis \cite{lee2018simple}      & 14.8 & 10.2 &  2.7 & 13.5 & 21.8 &  4.7 \\
SML \cite{jung2021standardized}            & 26.0 & 24.7 & 12.2 &  5.1 & 13.3 &  3.0 \\
JSRNet \cite{vojir2021road}                 & 20.2 & 29.3 & 13.7 & 18.6 & 24.5 & 11.0 \\
Mask2Former \cite{cheng2022masked}         & 25.2 & 18.2 & 15.3 &  5.0 & 21.9 &  4.8 \\
Max Entropy \cite{chan2021entropy} & 49.2 & 39.5 & 28.7 & 47.9 & 62.6 & 48.5 \\
SynBoost \cite{di2021pixel}                & 34.7 & 17.8 &  9.9 & 44.3 & 41.7 & 37.5 \\
Dense Hybrid \cite{grcic2022densehybrid}            & 54.1 & 24.1 & 31.0 & 45.7 & 50.1 & 50.7 \\
PEBEL \cite{tian2022pixel}                   & 38.8 & 27.2 & 14.4 & 29.9 &  7.5 &  5.5 \\
Rba \cite{nayal2023rba} & 55.7 & \textbf{52.1}  & 46.8 & 58.4 & 58.8 & 60.9 \\
Maskomaly \cite{ackermann2023maskomaly} & 55.4 & \underline{51.5} & \textbf{49.9} & - & - & - \\
Mask2Anomaly$^\dagger$ \cite{rai2023unmasking}                       & \textbf{60.4} & 45.7 & 48.6 & \textbf{61.4} &\underline{70.3} & \underline{69.8} \\
$\text{VL-Anomaly}$(ours) & \underline{59.6} & 50.1 & \underline{48.7} & \underline{61.2} & \textbf{73.4} & \textbf{70.1} \\
\bottomrule[1.5pt]
\multicolumn{7}{p{.96\linewidth}}{
            \textit{Note:} $\uparrow$ indicates higher is better. The best and second best results are \textbf{bold} and \underline{underlined}, respectively.
        }
\end{tabular}
\end{table}

\begin{table}[ht] 
\centering 
\caption{Ablation on pipeline design.}
\label{tab:ablation_incremental}
\scriptsize
\setlength{\tabcolsep}{5.pt} 
\begin{tabular}{l|cccc} 
\toprule[1.5pt] 
\textbf{Setting} & \textbf{$\text{FPR}_{95}$ $\downarrow$} & \textbf{AuPRC $\uparrow$} & \textbf{AuROC $\uparrow$} & \textbf{FPS $\uparrow$} 
\\ 
\midrule[1.pt] 
Baseline & 4.4 & 46.0 & 93.6 & \textbf{8.3}
\\ 
+ MLP \& Hand-crafted prompt  & 13.4 & 60.1 & 95.1 & 8.3
\\
$\rightarrow$ PL-Aligner \& S$_{\text{text}}$ & 9.2 & 64.0 & 95.1 & 7.9
\\ 
$\rightarrow$ Learnable prompt \cite{zhou2022coop} & 9.2 & \textbf{69.7} & 95.4 & 7.9
\\ 
+ S$_{\text{img}}$ (Full design) & \textbf{8.4} & 69.5 & \textbf{96.0} & 6.7 \\
\bottomrule[1.5pt]
\multicolumn{5}{p{.96\linewidth}}{
            \textit{Note:} Starting from the baseline, we progressively add prompts, PL-Aligner and multi-source inference components (S$_{\text{text}}$:text-guided similarity, S$_{\text{img}}$: CLIP-based image-text similarity). $\uparrow$ indicates higher is better, $\downarrow$ indicates lower is better.
        }
\end{tabular} 
\end{table}

\begin{table}[ht]
\centering
\caption{Ablation on pl-aligner layers.}
\label{tab:plaligner_ablation_layers}
\scriptsize
\setlength{\tabcolsep}{6.pt} 
\begin{tabular}{l|c|ccc}
\toprule[1.5pt]
\textbf{Alignment} & \textbf{Aligner} & \textbf{$\text{FPR}_{95}$ $\downarrow$} & \textbf{AuPRC $\uparrow$} & \textbf{AuROC $\uparrow$} \\
\midrule[1.pt]
Pixel-only  & MLP  & 13.4 & 60.1 & 93.6 \\
Pixel-only  & Cross-attn. \cite{ghiasi2022scaling} & 9.2 & 69.4 & 95.4 \\
Pixel-only       & PL-Aligner        & 9.1 & 69.4 & 95.6 \\
Mask-only        & PL-Aligner         & 16.7 & 55.4 & 78.6 \\
Pixel+Mask & PL-Aligner  & \textbf{8.4} & \textbf{69.5} & \textbf{96.0} \\
\bottomrule[1.5pt]
\multicolumn{5}{p{.96\linewidth}}{
            \textit{Note:} We compare aligning only with pixel-level, only with mask-level and using both layers (pixel+mask). $\uparrow$ indicates higher is better, $\downarrow$ indicates lower is better.
        }
\end{tabular}
\end{table}

\subsection{Evaluation Results}
\subsubsection{Results on RoadAnomaly and SMIYC-RA21/RO21}
Table~\ref{tab-ra-smiyc} summarizes the results on RoadAnomaly and the two subsets of the SMIYC benchmark, RA21 and RO21. VL-Anomaly consistently improves upon existing methods under the same Mask2Anomaly backbone. On RoadAnomaly, it achieves an AuROC of 96.8, which is +0.6 higher than Mask2Anomaly, and reduces $\text{FPR}_{95}$ from 13.2 to 12.9. On RA21, VL-Anomaly improves AuPRC by +6.4 compared with Mask2Anomaly. On RO21, it achieves the highest AuROC of 99.7, surpassing Mask2Anomaly by +0.6. When averaged across the three datasets, VL-Anomaly achieves the best overall performance. This aligns with our core idea, as the notable reduction in $\text{FPR}_{95}$ and the improvement in AuPRC primarily stem from fewer false positives in semantically normal background regions.

\subsubsection{Results on Fishyscapes}
The results on the Fishyscapes benchmark are reported in Table~\ref{tab-fs}. On the Static subset, VL-Anomaly achieves an AuROC of 98.7, outperforming Mask2Anomaly by +0.4, while maintaining a competitive $\text{FPR}_{95}$ of 2.3. On the more challenging Lost and Found subset, VL-Anomaly shows clear improvements, increasing AuPRC from 46.0 to 69.5 (+23.5) and raising AuROC from 93.6 to 96.0 (+2.4). These results confirm that VL-Anomaly not only excels in simpler settings but also generalizes well to complex scenarios.

\subsubsection{Results on SMIYC}
Table~\ref{tab-smiyc} presents the results on the RA21 and RO21 subsets of the SMIYC benchmark. Traditional confidence-based or distance-based approaches such as MSP, Mahalanobis and SML perform poorly, with $\text{F1}^*$ scores below 15 on RA21 and below 12 on RO21, showing their limitations in complex driving scenes. Mask-based segmentation frameworks such as Rba and Mask2Anomaly achieve notable improvements, with Mask2Anomaly reaching an sIoU of 60.4 on RA21 and an $\text{F1}^*$ of 69.8 on RO21. Building on this strong baseline, VL-Anomaly achieves further gains. On RA21, it achieves the best $\text{F1}^*$ of 48.7. On RO21, it improves PPV from 70.3 to 73.4 (+3.1) and $\text{F1}^*$ from 69.8 to 70.1 (+0.3), while maintaining a comparable sIoU of 61.2. Overall, VL-Anomaly delivers the best or near-best performance on both RA21 and RO21.

\subsection{Ablation Study}

All the ablation study results reported in this section are from the Fishyscapes lost and found validation dataset~\cite{blum2021fishyscapes}.

\begin{figure*}[t]
\centering
\includegraphics[width=\linewidth]{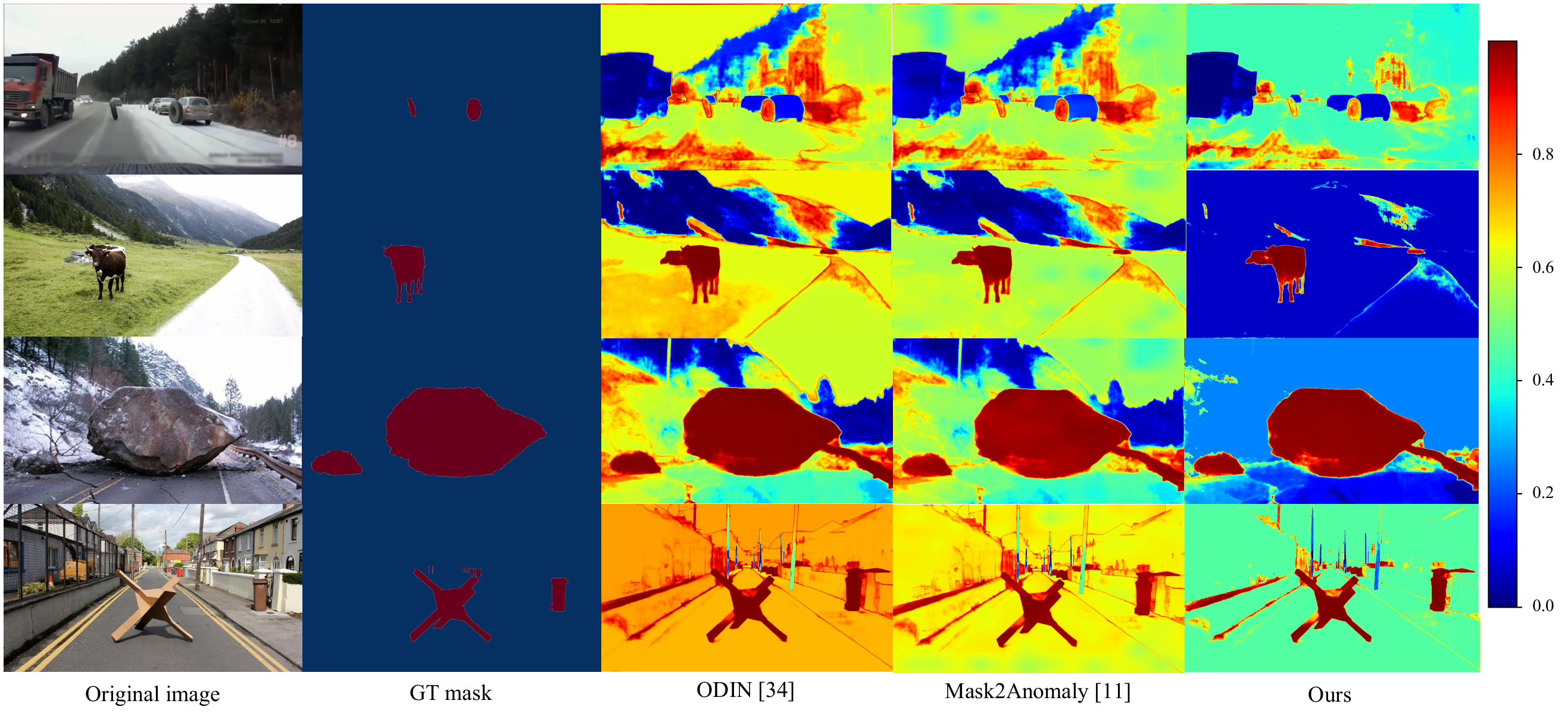}
\caption{Qualitative comparison of anomaly segmentation results on the Road Anomaly dataset~\cite{lis2019detecting}. We compare the outlier score maps predicted by our method with those generated by MSP~\cite{hendrycks2017a} and Mask2Anomaly~\cite{rai2023unmasking}, using the same backbone for a fair comparison. For visualization, all scores are normalized to the same range. Our method more effectively suppresses false positives in semantically normal background regions, while competing approaches often yield blurred or spurious activations in these areas.}
\label{fig:vis_result}
\end{figure*}

\begin{figure}[t]
\centering
\includegraphics[width=\linewidth]{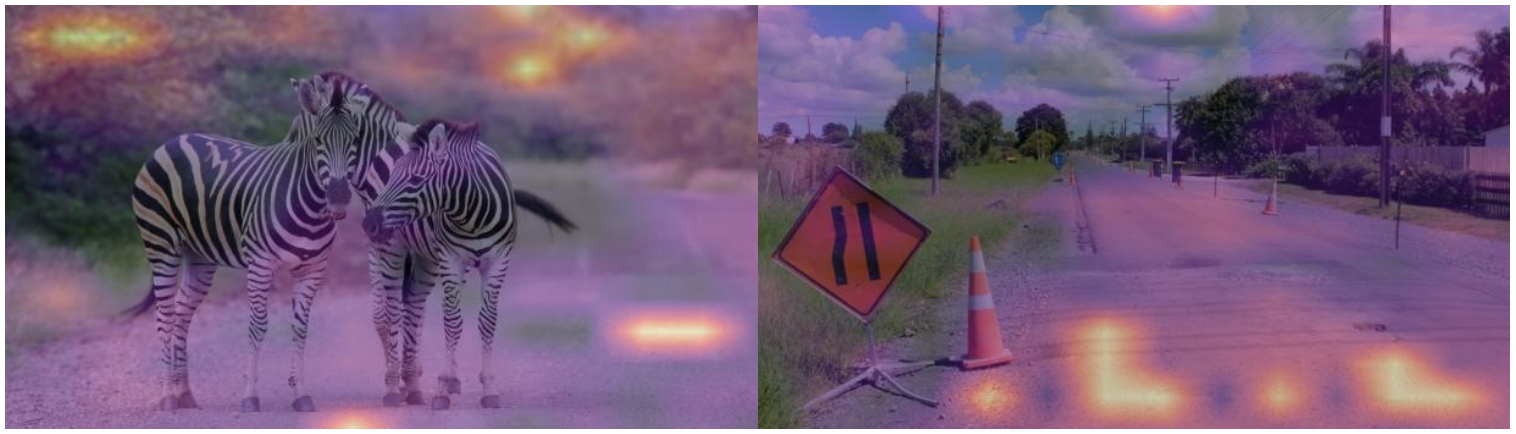}
\caption{Visualization of the similarity between image features and the constructed text prompts. The highlighted areas show where the model associates image regions with specific semantic categories, demonstrating the effectiveness of our prompt learning strategy in guiding cross-modal alignment.}
\label{fig:vis_analysis}
\end{figure}

\subsubsection{Pipeline design}
We conduct an ablation study in Table~\ref{tab:ablation_incremental} to evaluate the contributions of each module. Starting from the baseline, adding an MLP-based pixel-level alignment with $S_{\text{text}}$ improves detection performance, although $\text{FPR}{95}$ increases. Replacing the MLP with our PL-Aligner further enhances results and reduces $\text{FPR}{95}$. Introducing prompts and moving from hand-crafted to learnable ones brings consistent gains, with AuPRC reaching 69.7 while maintaining 7.9 fps. Finally, incorporating $S_{\text{img}}$ into the multi-source inference delivers the best overall performance at 6.7 fps, which represents only a modest decrease of 1.6 fps compared with the baseline, indicating that our innovations incur minimal efficiency cost.

\subsubsection{PL-Aligner layer design}
Table~\ref{tab:plaligner_ablation_layers} presents an ablation study on different PL-Aligner configurations. 
When alignment is performed only at the pixel level, simple MLP projection provides limited improvement, whereas adding cross-attention markedly reduces $\text{FPR}{95}$ and boosts AuPRC, confirming the necessity of modeling interactions between visual and textual features.
Our pixel-level PL-Aligner achieves comparable gains to cross-attention, but with slightly higher AuROC, showing its effectiveness as a lightweight alternative.
By contrast, alignment solely at the mask level performs poorly, indicating that high-level queries alone are insufficient to capture fine-grained semantic cues.
The full design, which integrates both pixel-level and mask-level alignment, achieves the best overall results, demonstrating that the two levels provide complementary information: pixel-level alignment anchors fine-grained semantics, while mask-level alignment reinforces structured category consistency.

\subsection{Qualitative Results}
Figure~\ref{fig:vis_result} shows anomaly maps from ODIN, Mask2Anomaly and VL-Anomaly.
Our method effectively suppresses false positives in semantically normal regions such as trees and vegetation, producing cleaner heatmaps with fewer spurious activations.
Figure~\ref{fig:vis_analysis} further illustrates the role of our prompt learning strategy. The highlighted areas correspond to normal background categories aligned with text prompts, while anomalous objects emerge as low-intensity regions.
This demonstrates that prompt-based cross-modal alignment introduces an additional semantic prior, enabling the model to better disentangle unexpected patterns from the background.

\section{Conclusion}
In this work, we presented VL-Anomaly, a vision–language guided anomaly segmentation framework that leverages semantic priors to enhance the segmentation of road anomalies. By introducing the PL-Aligner, we explicitly align visual features with CLIP text embeddings, effectively suppressing false positives in semantically normal regions. Furthermore, our multi-source inference strategy fuses detector confidence, text-guided similarity and CLIP-based Image Similarity to produce more reliable anomaly predictions. Extensive experiments on RoadAnomaly, SMIYC and Fishyscapes benchmarks demonstrate the superiority and generalization ability of our method over state-of-the-art baselines.
Although the results are promising, our multi-source inference strategy remains constrained by weights manually adjusted based on multiple datasets, which may limit scalability and automation. Adaptive or data-driven weight learning would offer a more principled alternative. Future work will focus on developing automatic weight optimization strategies to further improve robustness and usability across diverse application settings.

\bibliographystyle{IEEEtran}
\bibliography{IEEEfull}

\end{document}